\documentclass[10pt,times,mathptm,psfig,conference]{IEEEtran}

\usepackage{color}
\usepackage{cite}
\usepackage{amsmath}
\usepackage{graphicx}
\usepackage[font=footnotesize]{caption}
\usepackage{epstopdf}
\usepackage{tabularx}
\usepackage{booktabs} 
\usepackage[usenames,dvipsnames]{xcolor}
\usepackage{eso-pic} 
\usepackage{textcomp} 
\usepackage{pifont}
\usepackage{url}
\usepackage{multirow}

\usepackage[caption=false,font=footnotesize]{subfig}
\usepackage{fixltx2e}

\let\OLDthebibliography\thebibliography
\renewcommand\thebibliography[1]{
  \OLDthebibliography{#1}
  \setlength{\parskip}{0pt}
  \setlength{\itemsep}{0pt plus 0.3ex}
}

\ifCLASSINFOpdf
\else
\fi

\IEEEoverridecommandlockouts

\AddToShipoutPicture*{\small \raisebox{0.95cm}
{\hspace*{2.2cm}
\begin{minipage}[c]{0.95\textwidth}
\footnotesize
{\setlength{\baselineskip}{0.5\baselineskip}
\textcopyright 2020 IEEE. Personal use of this material is permitted.  Permission from IEEE must be obtained for all other uses, in any current or future media, including reprinting/republishing this material for advertising or promotional purposes, creating new collective works, for resale or redistribution to servers or lists, or reuse of any copyrighted component of this work in other works.
\par}
\end{minipage}}}%

\AddToShipoutPicture*{\small \raisebox{1.45cm}
{\hspace*{-17.35cm}
\begin{minipage}[c]{0.95\textwidth}
\footnotesize
{\setlength{\baselineskip}{0.5\baselineskip}
\hspace*{-2.5mm}This document is the paper as accepted for presentation at the IEEE International Symposium on Circuits and Systems (ISCAS) 2020.
\par}
\end{minipage}}}%

\begin{document}

\title{\vspace*{-5mm}A 28-nm Convolutional Neuromorphic Processor Enabling Online Learning with Spike-Based Retinas\vspace*{-9mm}}
\author{\IEEEauthorblockN{Charlotte Frenkel\thanks{This work was supported by the fonds europ\'een de d\'eveloppement r\'egional FEDER, the Wallonia within the ``Wallonie-2020.EU'' program, the Plan Marshall and the FRS-FNRS of Belgium. E-mail: charlotte.frenkel@uclouvain.be}, Jean-Didier Legat and David Bol}
\IEEEauthorblockA{ICTEAM Institute, Universit\'e catholique de Louvain, Louvain-la-Neuve, Belgium}\vspace*{-2mm}}

\maketitle

\begin{abstract} 
In an attempt to follow biological information representation and organization principles, the field of neuromorphic engineering is usually approached \textit{bottom-up}, from the biophysical models to large-scale integration \textit{in silico}. While ideal as experimentation platforms for cognitive computing and neuroscience, bottom-up neuromorphic processors have yet to demonstrate an efficiency advantage compared to specialized neural network accelerators for real-world problems. \textit{Top-down} approaches aim at answering this difficulty by (i) starting from the applicative problem and (ii) investigating how to make the associated algorithms hardware-efficient and biologically-plausible. In order to leverage the data sparsity of spike-based neuromorphic retinas for adaptive edge computing and vision applications, we follow a top-down approach and propose SPOON, a 28-nm event-driven CNN (eCNN). It embeds online learning with only 16.8-\% power and 11.8-\% area overheads with the biologically-plausible direct random target projection (DRTP) algorithm. With an energy per classification of 313nJ at 0.6V and a 0.32-mm$^2$~area for accuracies of 95.3\% (on-chip training) and 97.5\% (off-chip training) on MNIST, we demonstrate that SPOON reaches the efficiency of conventional machine learning accelerators while embedding on-chip learning and being compatible with event-based sensors, a point that we further emphasize with N-MNIST benchmarking.
\end{abstract}

\vspace*{1mm}
\section{Introduction}
\vspace*{2mm}

\begin{figure}[!t]
\centering
\noindent\includegraphics[width=0.61\columnwidth]{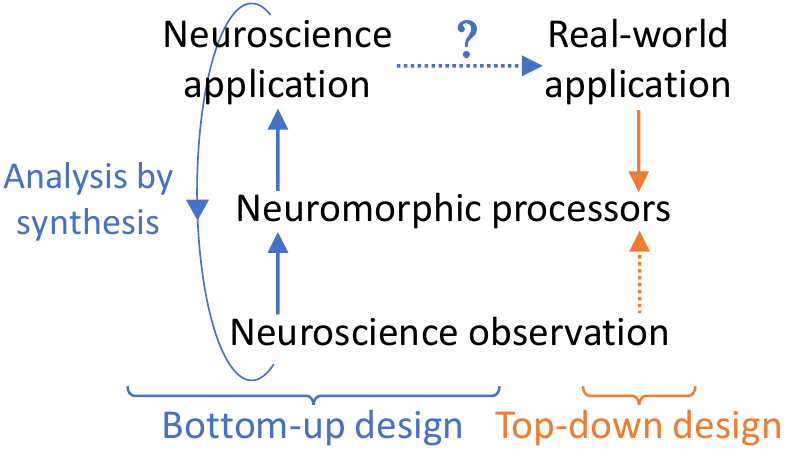}
\vspace*{-2mm}
\caption{Summary of bottom-up and top-down neuromorphic design approaches.}
\vspace*{-1mm}
\label{fig_motivation}
\end{figure}

The field of neuromorphic engineering takes its roots into the discovery that the MOS transistor operated in subtreshold could directly emulate the ion channels dynamics in the brain~\cite{Mead89}. This led to a long tradition of \textit{bottom-up} design since the late 1980s, going from neuroscience observation and biophysical neuron and synapse models to analog and digital small-scale~\cite{Qiao15,Moradi18,Mayr16,Seo11,Frenkel19a,Frenkel19b} and large-scale~\cite{Schemmel10,Benjamin14,Painkras13,Akopyan15,Davies18} integrations \textit{in silico}. Bottom-up neuromorphic processors are thus ideal as experimentation platforms for cognitive computing and neuroscience~\cite{Cauwenberghs13,Indiveri15}, for which they even help to reverse-engineer the brain with \textit{analysis by synthesis}~(Fig.~\ref{fig_motivation}). However, the key challenge lies in applying them to real-world scenarios, which is yet to be demonstrated with an efficiency advantage compared to conventional frame-based artificial and convolutional neural network (ANN, CNN)~\mbox{hardware accelerators~\cite{Indiveri15,Park19}.}

In order to address the difficulty of bottom-up neuromorphic designs in tackling real-world problems efficiently, a few \textit{top-down} designs have recently been proposed for \textit{adaptive} edge computing (e.g.,~\cite{Kim15, Buhler17, Park19, Chen18}), ensuring both robustness to uncontrolled environments and low-cost deployment for applications power- and resource-constrained during the training phase. Starting from this applicative problem, top-down designs investigate how to embed online learning with a focus on hardware efficiency and biological plausibility~(Fig.~\ref{fig_motivation}). However, top-down designs currently appear to be either spiking neural networks (SNNs) with event-driven processing at the expense of accuracy~\cite{Kim15, Buhler17} or binary neural networks (BNNs) with high accuracy at the expense of conventional frame-based processing~\cite{Park19}. The chip from Chen~\textit{et~al.}~\cite{Chen18} allows exploring both sides with an SNN embedding STDP that can also be programmed~\mbox{as a BNN using offline-trained weights.}

\begin{figure}[!t]
\centering
\vspace*{-2.0mm}%
\noindent\includegraphics[width=1.0\columnwidth]{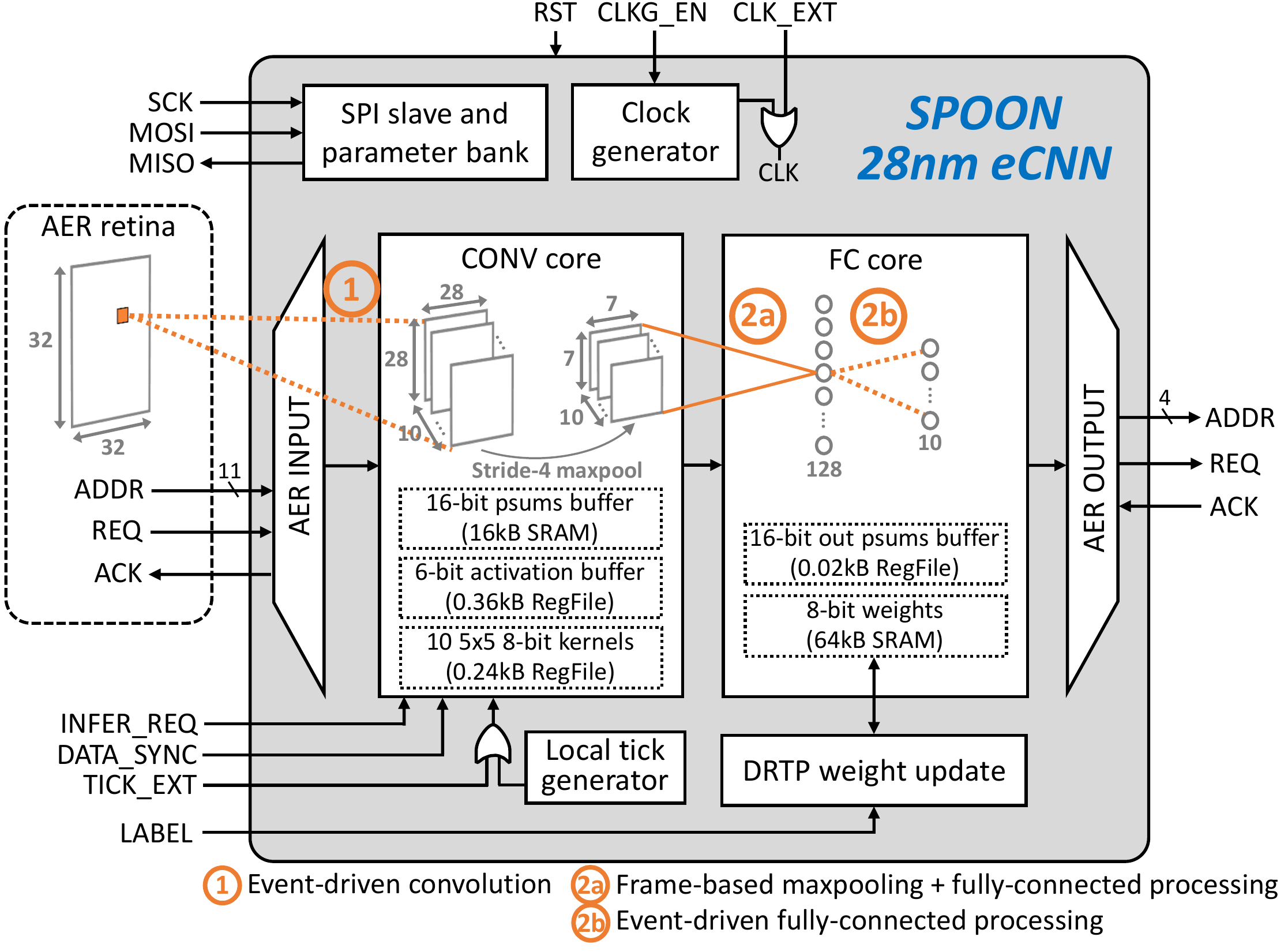}
\caption{Block diagram of the SPOON event-driven CNN (eCNN) processor.}%
\vspace*{-1mm}%
\label{fig_architecture}
\end{figure}

\begin{figure*}[!t]
\centering
\vspace*{-5mm}%
\noindent\includegraphics[width=1.0\textwidth]{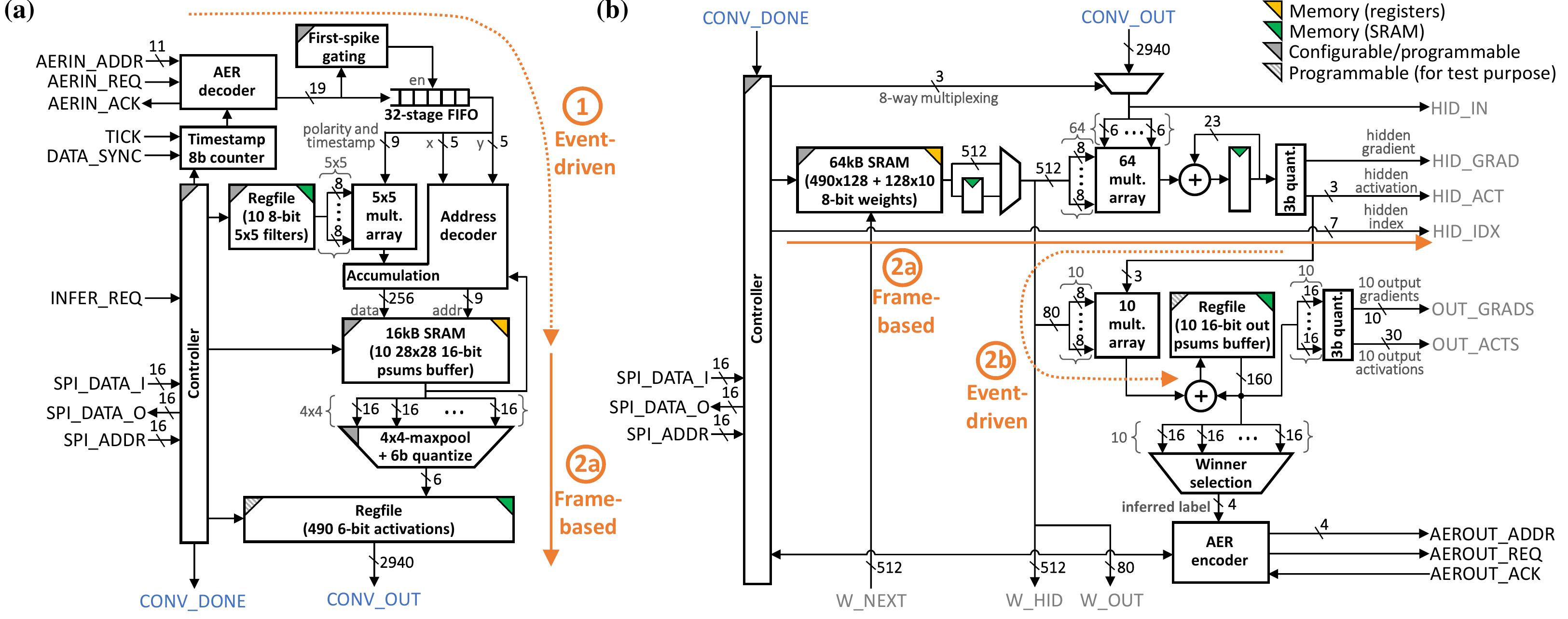}
\vspace*{-5mm}%
\caption{Circuit architecture of (a) the CONV core and of (b) the FC core. Detailed control signals, data gating and overflow protection mechanisms are not shown for clarity. Event-driven and frame-driven processing are highlighted following the conventions of Fig.~\ref{fig_architecture}.\vspace*{-3mm}}
\label{fig_circuits_cores}
\end{figure*}

Therefore, in this work, we propose SPOON (standing for \underline{\smash{sp}}iking \underline{\smash{o}}nline-learning c\underline{\smash{o}}nvolutional \underline{\smash{n}}euromorphic processor), an event-driven CNN (eCNN) for adaptive edge computing. Event-driven convolutions with time-to-first-spike coding leverage sparsity from event-based neuromorphic retinas~\cite{Orchard15a}, an idea now also explored for conventional machine learning accelerators~\cite{Goetschalckx19}, while a combination with frame-based processing ensures maximum data reuse and parallelism in fully-connected layers. SPOON embeds online learning at low power and area overheads with the biologically-plausible direct random target projection (DRTP) algorithm~\cite{Frenkel19c}, which we introduced recently to release the key issues of the successful error backpropagation (BP) algorithm~\cite{Rumelhart86} precluding hardware efficiency and biological plausibility. We demonstrate that, to the best of our knowledge, only SPOON allows reaching the efficiency of conventional machine learning accelerators while embedding on-chip learning and being compatible with event-based sensors. The architecture of SPOON is described in Section~\ref{sec_arch}, implementation details and benchmarking results on MNIST and N-MNIST are given in Section~\ref{sec_results}.

\newpage

\vspace*{-10mm}
\section{Architecture} \label{sec_arch}
\vspace*{2mm}

A block diagram of SPOON is shown in Fig.~\ref{fig_architecture}. Four-phase-handshake address-event representation (AER) buses~\cite{Boahen00} are used for event-driven handling of input sensor spikes and of output inferences. All weights and parameters can be programmed and readback with an SPI bus. As neuromorphic vision sensors send spikes encoding temporal contrast~\cite{Vanarse16}, pixels with the highest luminosity change spike first with \texttt{ON} (positive delta) or \texttt{OFF} (negative delta) events, conveying useful data for edge detection. In order to efficiently extract this information, we use time-to-first-spike encoding (i.e. timing code)~\cite{Thorpe01} in the convolutional layers, which are handled in the CONV core (Section~\ref{ssec_conv}). In order to match the time constant of SPOON with the given application, time ticks can be retrieved either from an external reference pin \texttt{TICK\_EXT} or generated internally by a configurable synchronous on-chip tick generator. Fully-connected layers are handled in the FC core (Section~\ref{ssec_fc}), which uses a combination of frame-based and event-driven processing for maximum data reuse and efficient handling of DRTP updates (Section~\ref{ssec_drtp}).

\begin{figure}[!t]
\centering
\vspace*{-3mm}%
\noindent\includegraphics[width=0.965\columnwidth]{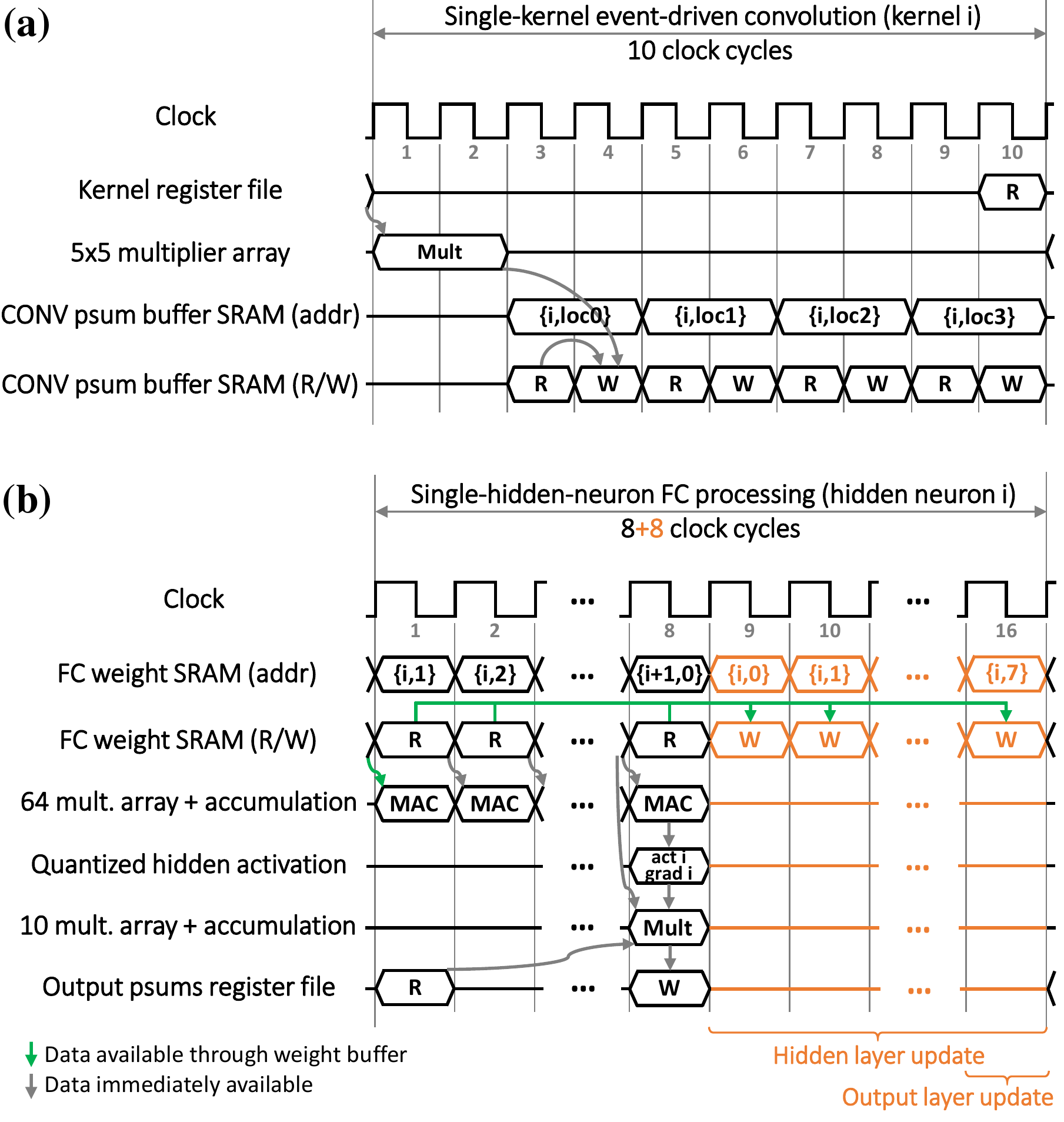}
\vspace*{-2mm}%
\caption{Timing diagrams for (a)~event-driven convolution in the CONV core and (b)~combination of hidden and output layer processing in the FC core. Black: cycles required for inference. Orange: optional DRTP update cycles.\vspace*{-3mm}}
\label{fig_multiplexing}
\end{figure}

\vspace*{1mm}
\subsection{Convolutional (CONV) core} \label{ssec_conv}
\vspace*{2mm}

The CONV core consists of a convolutional layer with 10 5$\times$5 8-bit programmable kernels followed by a stride-4 maxpooling layer, kernels are randomly initialized upon reset. The circuit architecture is shown in Fig.~\ref{fig_circuits_cores}(a). Following the dataflow highlighted in Fig.~\ref{fig_architecture}, convolutions are carried out in an event-driven fashion, while frame-based maxpooling is triggered before processing the fully-connected layers. Input AER events from the sensor are encoded with an 11-bit address, which covers the pixel \{$x$,$y$\} coordinates for 32$\times$32 images and an \texttt{ON}/\texttt{OFF} polarity bit. Based on the \texttt{TICK} time reference and the \texttt{DATA\_SYNC} pin that signals the start of an input sample, input events are concatenated with an 8-bit timestamp before being pushed into a 32-stage FIFO.
 
Event-driven convolutions follow the timing diagram of Fig.~\ref{fig_multiplexing}(a), where the 10 kernels are processed sequentially. The 9-bit timestamp, including the input event polarity bit, is first multiplied with all values in the current kernel $i$ in a 5$\times$5 multiplier array. The partial sums (psums) of the feature map elements associated to kernel $i$ and input pixel \{$x$,$y$\} coordinates, stored in a 16-kB SRAM, are then updated. Due to SRAM aspect ratio constraints, this update is split in four 256-bit read/write accesses whose locations are given by an address decoder. An overflow protection mechanism emulates a hardtanh activation function. Maxpooling is automatically carried out after the FIFO has been emptied and the 8-bit timestamp counter falls to zero. It is followed by a quantization to 6 bits with configurable rescaling. The CONV core then outputs 490 6-bit activations (\texttt{CONV\_OUT}) and a \texttt{CONV\_DONE} trigger to enable the FC core, which is clock-gated otherwise.

Depending on the event-driven sensor use case, two features can be used to adjust the accuracy-energy tradeoff. First, the \textit{first-spike gating} block can be enabled to keep only the most-informative first spike of each pixel, thus dropping subsequent events. Second, the \texttt{INFER\_REQ} pin can be used to request inference at any time by triggering maxpooling before the 8-bit timestamp counter falls to zero, thus ignoring all subsequent less-informative events. 

\vspace*{1mm}
\subsection{Fully-connected (FC) core} \label{ssec_fc}
\vspace*{2mm}

\begin{figure}[!t]
\centering
\vspace*{-5mm}
\noindent\includegraphics[width=0.57\columnwidth]{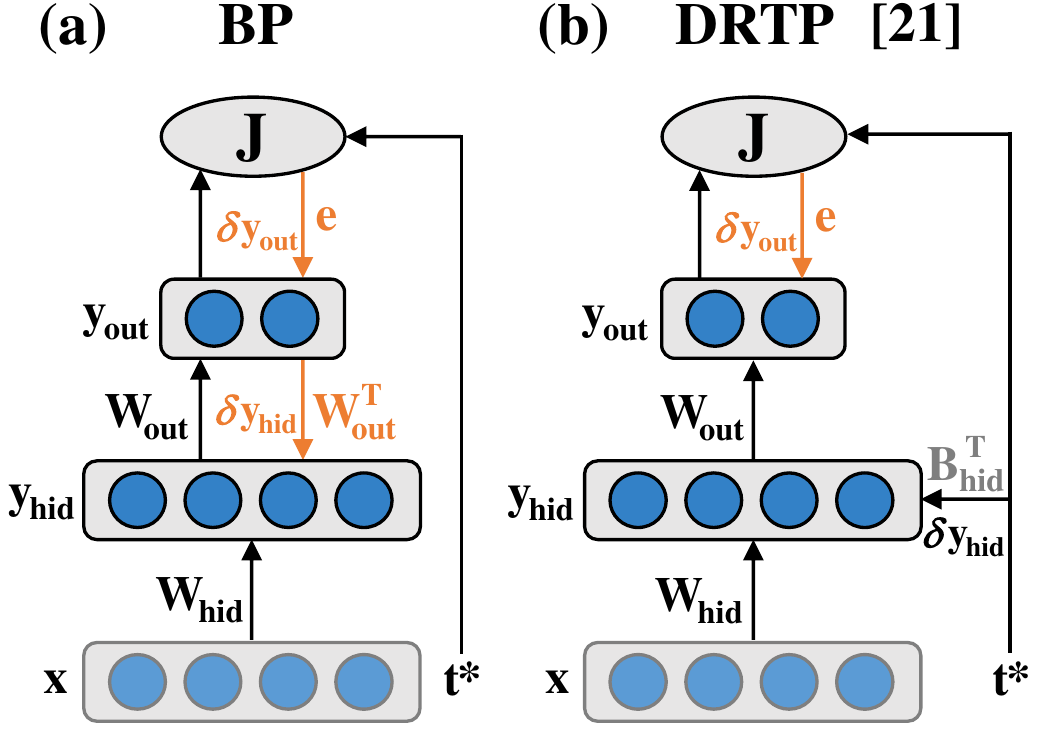}
\vspace*{-1.5mm}
\caption{(a) Backpropagation of error (BP) algorithm~\cite{Rumelhart86}. (b) Direct random target projection (DRTP) algorithm~\cite{Frenkel19c}. Adapted from~\cite{Frenkel19c} for the case of a single hidden layer, matching the conventions of Figs.~\ref{fig_circuits_cores} and~\ref{fig_circuits_drtp}.\vspace*{-2mm}}
\label{fig_drtp}
\end{figure}

The FC core consists of a 128-neuron fully-connected hidden layer followed by a 10-neuron output layer, both with 8-bit programmable weights that are automatically initialized to zero for online learning (Section~\ref{ssec_drtp}). The circuit architecture and the associated timing diagram are shown in Figs.~\ref{fig_circuits_cores}(b) and~\ref{fig_multiplexing}(b), respectively. As highlighted in Fig.~\ref{fig_architecture}, the hidden layer output $y_{hid} = f_{hid}(W_{hid} \hspace*{0.3mm} x)$ is computed with a conventional frame-based approach as all the inputs are immediately available when receiving the \texttt{CONV\_DONE} trigger, where $W_{hid}$ represents the hidden layer weights, $f_{hid}$ is the hidden layer activation function and $x$ is the input from the CONV core~(Fig.~\ref{fig_drtp}). The hidden neurons are evaluated sequentially and inputs are processed by batch of 64. It requires 8 cycles to retrieve the 500 weights associated to a hidden neuron, including both the 490 weights $W_{hid,i}$ connecting to the inputs and the 10 weights $W_{out,i}$ connecting to the output layer neurons, where the index $i$ denotes hidden neuron $i$. Once the weighted sum of inputs $W_{hid,i} \hspace*{0.3mm} x$ of hidden neuron $i$ has been computed, output layer processing is triggered in an event-driven fashion to ensure maximum data reuse:%
\begin{itemize}
\item[--] the activation $y_{hid,i}$ is obtained by quantizing $W_{hid,i} \hspace*{0.3mm} x$ to 3 bits with a hardtanh function, whose binary derivative is one in the linear range and zero elsewhere (\texttt{HID\_ACT} and \texttt{HID\_GRAD} in Fig.~\ref{fig_circuits_cores}(b)),
\item[--] if the derivative is non-zero, DRTP updates can be directly applied to $W_{hid,i}$ (Fig.~\ref{fig_multiplexing}, orange), otherwise they are skipped (Section~\ref{ssec_drtp}),
\item[--] $W_{out,i} \hspace*{0.3mm} y_{hid,i}$ is added to the 10 output psums,
\item[--] as the final activation and derivative of the output neurons are not yet available for the current sample, a DRTP update of $W_{out,i}$ is triggered based on buffered previous sample data (Section~\ref{ssec_drtp}).
\end{itemize}
Finally, when all the hidden neurons have been processed, the output layer activation $y_{out} = f_{out}(W_{out} \hspace*{0.3mm} y_{hid})$ is obtained by quantizing the output psums to 3 bits with a hardsigmoid function, whose binary derivative is one in the linear range and zero elsewhere (\texttt{OUT\_ACTS} and \texttt{OUT\_GRADS} in Fig.~\ref{fig_circuits_cores}(b)).

\vspace*{1mm}
\subsection{On-chip online training with direct random target projection (DRTP)} \label{ssec_drtp}
\vspace*{2mm}

Building on feedback alignment techniques~\cite{Lillicrap16,Nokland16}, which were proposed to solve the \textit{weight transport problem} of BP~\cite{Rumelhart86} (i.e.~requirement for weight symmetry in the forward and backward pathways), we proposed in~\cite{Frenkel19c} the direct random target projection (DRTP) algorithm to release not only the weight transport problem, but also \textit{update locking} (i.e.~requirement for full forward and backward passes before the weights can be updated). As these are the two key issues that preclude BP from being hardware-efficient and biologically plausible~\cite{Frenkel19c}, DRTP is a low-cost algorithm suitable for deployment at the edge. It relies only on feedforward and local computation (Fig.~\ref{fig_drtp}) and estimates the hidden layer loss gradient $\delta y_{hid}$ as a projection of the target vector $t^*$ (i.e. one-hot encoded labels) with a fixed random matrix $B_{hid}$. This operation corresponds to a simple label-dependent random vector selection, which can be quantized down to binary resolutions with only a negligible impact on DRTP performance. The hidden layer weight updates~\mbox{$\Delta W_{hid}$ can then be computed as}%

\vspace*{-3mm}%
\begin{equation}\label{eq_drtp}
\Delta W_{hid} = -\eta_{hid} \left( \delta y_{hid} \odot f_{hid}'(W_{hid} x) \right) x^T,
\end{equation}
\vspace*{-5mm}%

\noindent where $\eta_{hid}$ is the hidden layer learning rate and $\odot$ denotes element-wise multiplication. As opposed to BP, $\delta y_{hid}$ is always non-zero in DRTP, the weights can thus be initialized to zero without precluding training convergence. The DRTP output layer update is identical to the BP update, i.e.

\vspace*{-3mm}%
\begin{equation}\label{eq_bp}
\Delta W_{out} = -\eta_{out} \left(e \odot f_{out}'(W_{out} y_{hid}) \right) y_{hid}^T,
\end{equation}
\vspace*{-5mm}%

\noindent where $\eta_{out}$ is the output layer learning rate.

\begin{figure}[!t]
\centering
\vspace*{-5mm}
\noindent\includegraphics[width=1.0\columnwidth]{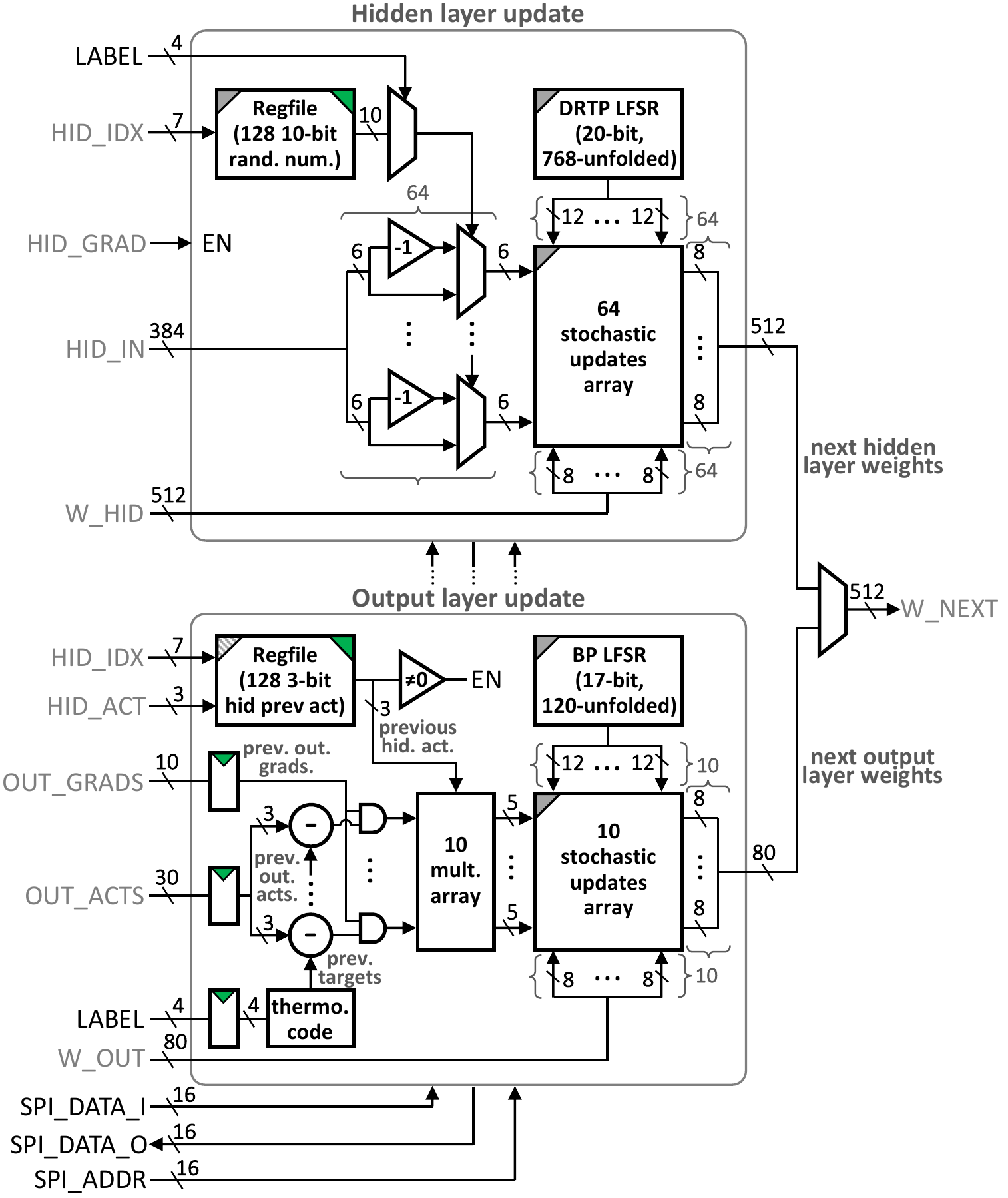}
\caption{Circuit architecture for the DRTP weight update module.\vspace*{-2mm}}
\label{fig_circuits_drtp}
\end{figure}

The circuit architecture for the DRTP weight update \mbox{module} of SPOON is shown in Fig.~\ref{fig_circuits_drtp}. According to Eq.~(\ref{eq_drtp}), the derivative $f_{hid}'$ (\texttt{HID\_GRAD}) of hidden neuron $i$, taking values $0$ or $1$, can be used as an enable signal for the hidden layer update module (Section~\ref{ssec_fc}). The fixed random binary matrix $B_{hid}$ is stored in a register file. A specific bit is selected based on the current hidden layer neuron index (\texttt{HID\_IDX}) and the training sample label (\texttt{LABEL}). Therefore, the only required computation is a label-dependent sign inversion to the inputs from the CONV core (\texttt{HID\_IN}), processed by batch of 64.~The obtained values are then used as probabilities conditioning random increments/decrements to the hidden layer weights $W_{hid,i}$ (\texttt{W\_HID}), depending on the values generated by a linear feedback shift register (LFSR) and a configurable learning rate. In order to parallelize the generation of 64 12-bit seeds with a single LFSR, we applied the unfolding technique~\cite{Parhi99}, similarly to the stochastic update mechanism that we proposed for the MorphIC SNN in~\cite{Frenkel19b}.

The output layer update follows Eq.~(\ref{eq_bp}), whose terms are buffered so that updates from the previous sample to output layer weights $W_{out,i}$ (\texttt{W\_OUT}) can be applied concurrently with the hidden layer updates of the current sample. The error $e$ is computed based on the previous label and output activations (\texttt{OUT\_ACTS}), where the binary output derivatives (\texttt{OUT\_GRADS}) act as a gating signal. If the previous hidden neuron activation is zero, updates are skipped. Otherwise, it is multiplied with the error and used in the stochastic updates array, which operates as in the hidden layer update module.

\begin{figure}[!t]
\centering
\vspace*{-5mm}
\noindent\includegraphics[width=0.9\columnwidth]{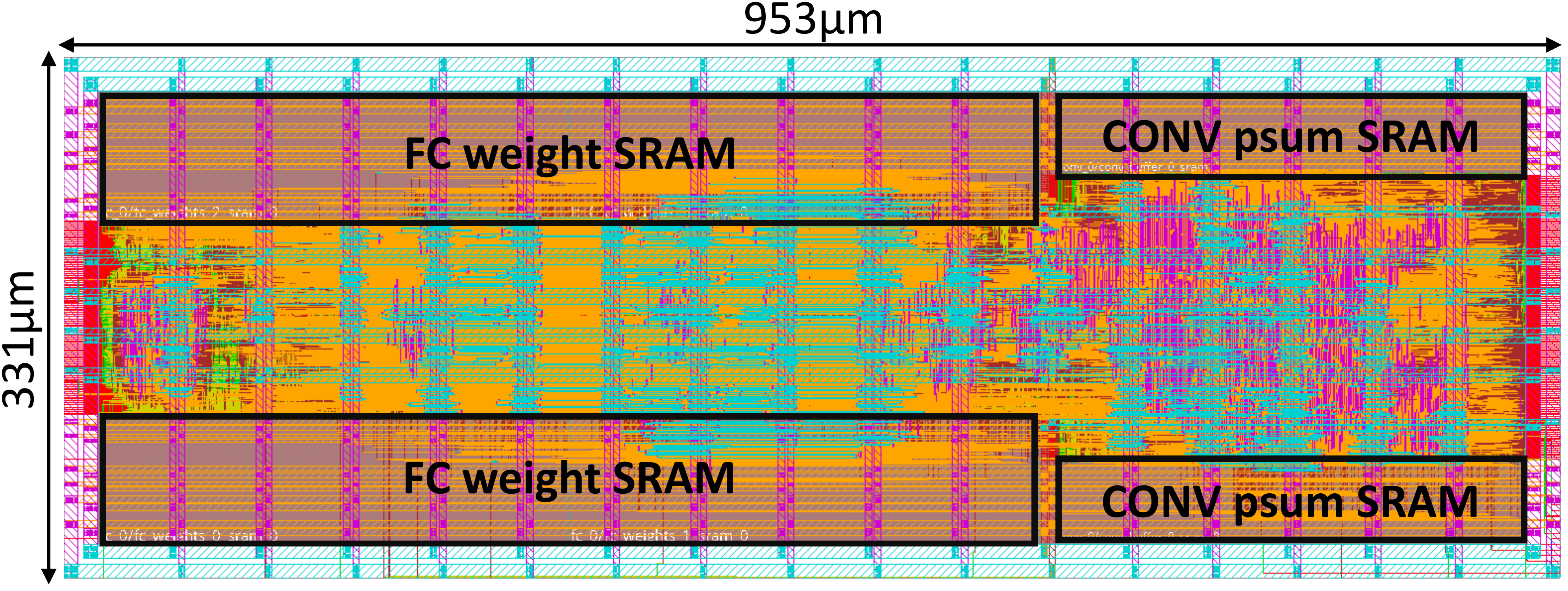}
\setcounter{figure}{6}%
\vspace*{-3.5mm}
\caption{SPOON layout view obtained under Cadence Innovus.\vspace*{-6mm}}
\label{fig_floorplan}
\end{figure}

\begin{table}[!t]
\caption{Specifications and pre-silicon performance metrics of SPOON.\vspace*{-1.5mm}}
\label{table_specs}
\renewcommand{\arraystretch}{0.9}
\centering
\resizebox{0.8\columnwidth}{!}{
\begin{tabular}{lc}
\toprule%
Technology & 28nm FDSOI CMOS \\
Implementation & Digital \\
Area & 0.32mm$^2$ (0.26mm$^2$ excl. rails)\\%
Topology & C5$\times$5@10--FC128--FC10 \\
Online learning & Stochastic DRTP, 8-bit weights \\
Time constant & Biological to accelerated \\
Supply voltage & 0.6V -- 1.0V \\
Max. clock frequency & 150MHz \\
Leakage power & 61$\mu$W at 0.6V \\
Energy for CONV core & 1.7nJ/event at 0.6V \\
Energy for FC core & 55nJ/inference at 0.6V \\
Online learning overhead & 16.8\% in power, 11.8\% in area\\
\bottomrule%
\end{tabular}}
\end{table}

\vspace*{1mm}
\section{Implementation and Benchmarking Results} \label{sec_results}
\vspace*{2mm}

SPOON has been taped out in a 28-nm FDSOI CMOS process, the layout is presented in Fig.~\ref{fig_floorplan}, while the specifications and pre-silicon performance metrics are reported in Table~\ref{table_specs}. It occupies an area of only 0.32mm$^2$. At 0.6V, SPOON has a leakage power of 61$\mu$W and the convolutional layer update consumes 1.7nJ per event in the CONV core. When convolution is over, the FC core requires 55nJ to update the fully-connected layers and to send the inferred label with the output AER bus. The high suitability of DRTP for adaptive edge computing is highlighted by its low power and area overheads of 16.8\% and 11.8\% compared to a design without online learning, respectively. In order to accelerate benchmarking, the accuracy results in the subsequent text were retrieved from an~\mbox{FPGA implementation of SPOON.}

\begin{figure}[!t]%
\centering
\vspace*{-5.5mm}%
\includegraphics[width=0.897\columnwidth]{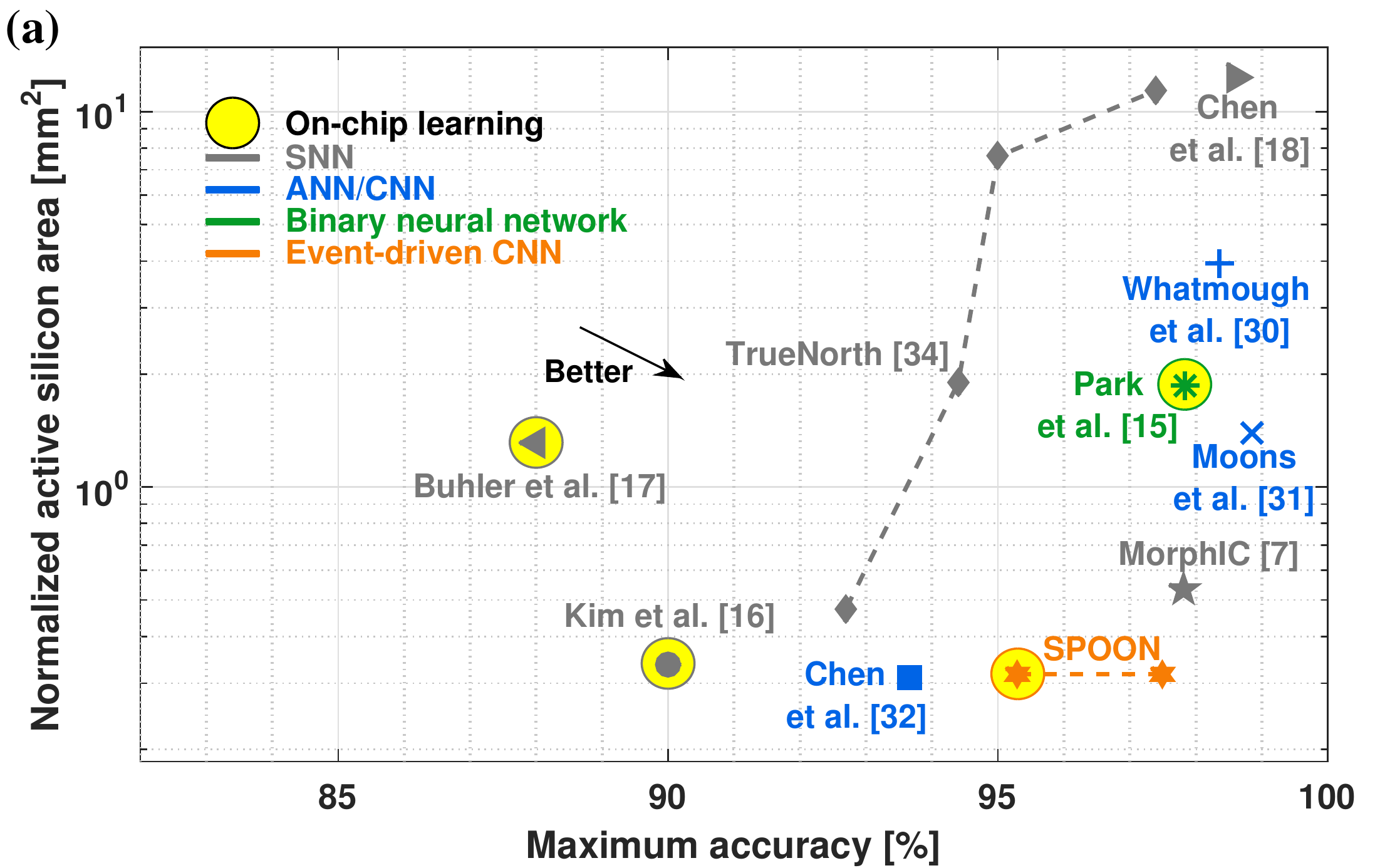}\\%
\includegraphics[width=0.897\columnwidth]{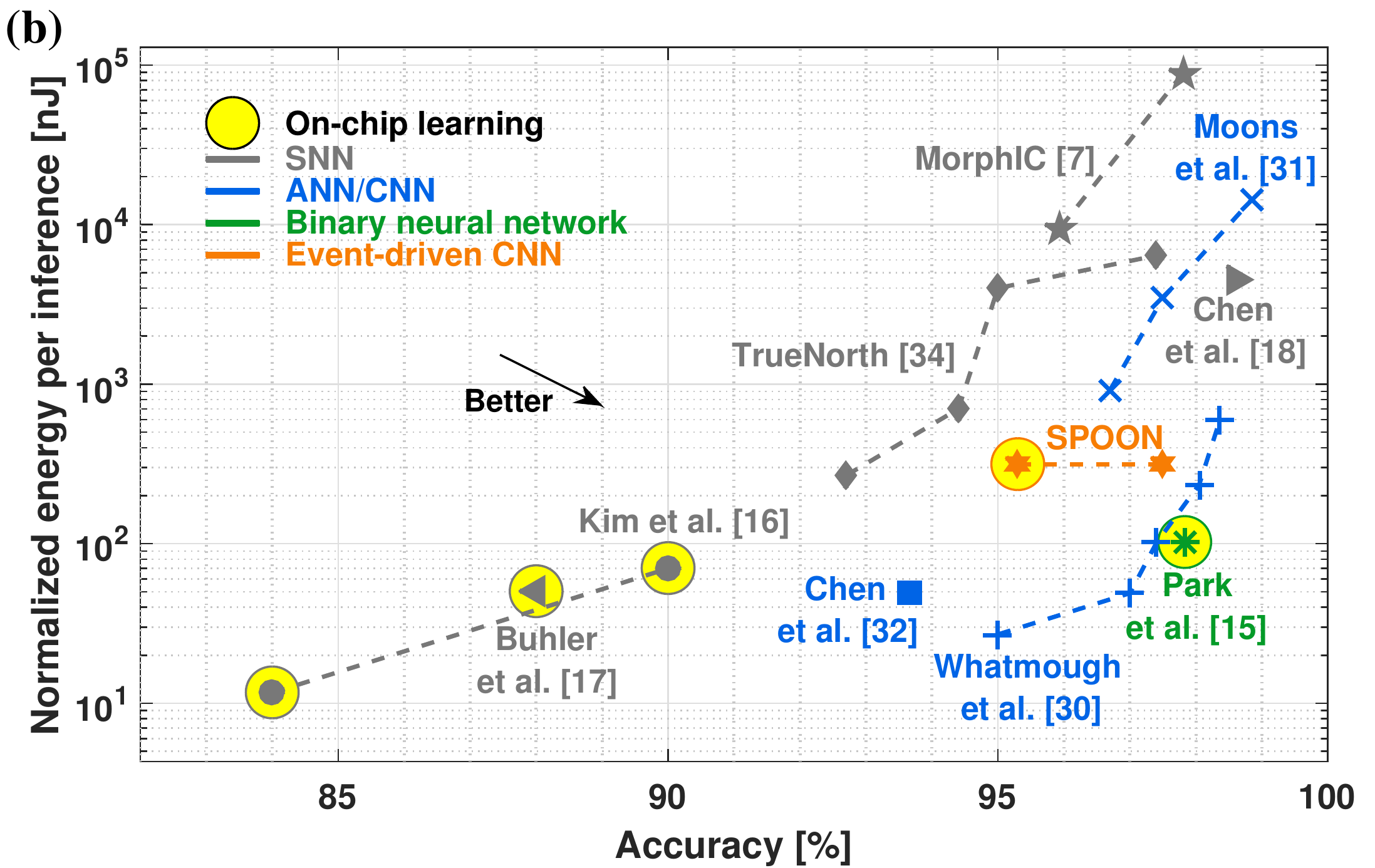}%
\caption{Accuracy-area-energy tradeoff normalized to 28nm for SNN, ANN, CNN and BNN accelerators on MNIST. Normalization has been carried out using the node factor, except for the 10-nm FinFET node from Chen \textit{et~al.}~\cite{Chen18} where data from~\cite{Mistry17} was used. Being a mixed-signal design, the chip from Buhler \textit{et~al.}~\cite{Buhler17} was not normalized. The non-preprocessed MNIST experiments reported for Chen \textit{et~al.}~\cite{Chen18} rely on offline BP-based BNN training. MNIST results for TrueNorth~\cite{Akopyan15} are reported in~\cite{Esser15}.\vspace*{-2.5mm}}
\label{fig_ppa}
\end{figure}

The accuracy-area-energy tradeoff on the MNIST dataset of handwritten digits~\cite{LeCun98} is shown in Fig.~\ref{fig_ppa} for conventional ANN and CNN machine learning accelerators~\cite{Whatmough17,Moons18,Chen19}, the BNN from Park~\textit{et~al.}~\cite{Park19}, the SNN offline-trained as a BNN from Chen~\textit{et~al.}~\cite{Chen18}, SNNs~\cite{Kim15,Buhler17,Akopyan15,Frenkel19b} and SPOON, which requires only 313nJ and 117$\mu$s per inference for an area of 0.32mm$^2$ using a time-to-first-spike encoding. Training SPOON for MNIST using an off-chip optimizer based on PyTorch~\cite{Paszke17} with quantization-aware training~\cite{Hubara17}, we reach a test-set accuracy of 97.5\%. When enabling on-chip DRTP-based online learning, where the convolution kernels are initialized and fixed to random values and plastic fully-connected weights are initialized to zero, SPOON reaches a test-set accuracy of 92.8\% after one epoch and 95.3\% after 100 epochs on the 60k-sample training set. It appears from Fig.~\ref{fig_ppa} that only SPOON reaches the efficiency of conventional machine learning accelerators while being compatible with event-based sensors and embedding on-chip online learning. 

The neuromorphic MNIST (N-MNIST) dataset~\cite{Orchard15b} is a spiking version of MNIST generated by an ATIS silicon retina~\cite{Posch11} mounted on a pan-tilt unit and moved in three saccades. As each active pixel spikes in average 4.8 times per saccade, leading to redundant information in time-to-first-spike coding, we use the single-spike-per-pixel mode of SPOON. Using only the first saccade of each sample, SPOON reaches a test-set accuracy of 93.8\% with offline-trained weights and of 90.2\% (one epoch) or 93.0\% (100 epochs) using on-chip online training,~\mbox{while consuming 665nJ per inference.}%

\vspace*{1mm}
\section{Conclusion} \label{sec_conclusion}
\vspace*{2mm}

In this paper, we presented the SPOON event-driven CNN, following a top-down neuromorphic design approach. We demonstrate that combining event-driven and frame-based processing with weight-transport-free update-unlocked training supports low-cost adaptive edge computing. Indeed, SPOON has an accuracy-area-energy tradeoff superior to SNNs and comparable to conventional machine learning accelerators while enabling online learning with spike-based sensors.


\newpage
\begin{thebibliography}{1}
\vspace*{2mm}

\bibitem{Mead89}
C. Mead, \textit{Analog VLSI and Neural Systems.} Reading, MA: Addison-Wesley, 1989.
\bibitem{Qiao15}
N. Qiao et al., ``A reconfigurable on-line learning spiking neuromorphic processor comprising 256 neurons and 128K synapses,'' \textit{Frontiers in Neuroscience}, vol.~9, no.~141, 2015.%
ù\vspace*{0.3mm}
\bibitem{Moradi18}
S. Moradi et al., ``A scalable multicore architecture with heterogeneous memory structures for Dynamic Neuromorphic Asynchronous Processors (DYNAPs),'' \textit{IEEE Transactions on Biomedical Circuits and Systems}, vol.~12, no. 1, pp. 106-122, 2018.
\bibitem{Mayr16}
C. Mayr et al., ``A biological-realtime neuromorphic system in 28 nm CMOS using low-leakage switched capacitor circuits,'' \textit{IEEE Trans. on Biomedical Circuits and Systems}, vol.~10, no.~1, pp.~243-254, 2016.%
\bibitem{Seo11}
J.-S. Seo et al., ``A 45nm CMOS neuromorphic chip with a scalable architecture for learning in networks of spiking neurons,'' \textit{Proc. of IEEE Custom Integrated Circuits Conference (CICC)}, 2011. 
\bibitem{Frenkel19a}
C. Frenkel et al., ``A 0.086-mm$^2$ 12.7-pJ/SOP 64k-synapse 256-neuron online-learning digital spiking neuromorphic processor in 28-nm CMOS,'' \textit{IEEE Transactions on Biomedical Circuits and Systems}, vol.~13, no.~1, pp.~145-158, 2019.
\bibitem{Frenkel19b}
C. Frenkel, J.-D. Legat and D. Bol, ``MorphIC: A 65-nm 738k-synapse/mm$^2$ quad-core binary-weight digital neuromorphic processor with stochastic spike-driven online learning'' \textit{IEEE Transactions on Biomedical Circuits and Systems}, vol.~13, no.~5, pp.~999-1010, 2019.
\bibitem{Schemmel10}
J. Schemmel et al., ``A wafer-scale neuromorphic hardware system for large-scale neural modeling,'' \textit{Proc. of IEEE International Symposium on Circuits and Systems (ISCAS)}, pp.~1947-1950, 2010.%
\bibitem{Benjamin14}
B. V. Benjamin et al., ``Neurogrid: A mixed-analog-digital multichip system for large-scale neural simulations,'' \textit{Proceedings of the IEEE}, vol.~102, no.~5, pp.~699-716, 2014.%
\bibitem{Painkras13}
E. Painkras et al., ``SpiNNaker: A 1-W 18-core system-on-chip for massively-parallel neural network simulation,'' \textit{IEEE Journal of Solid-State Circuits}, vol. 48, no. 8, pp. 1943-1953, 2013.
\bibitem{Akopyan15}
F. Akopyan et al., ``TrueNorth: Design and tool flow of a 65 mW 1 million neuron programmable neurosynaptic chip,'' \textit{IEEE Transactions on Computer-Aided Design of Integrated Circuits and Systems}, vol.~34, no.~10, pp.~1537-1557, 2015.%
\bibitem{Davies18}
M. Davies et al., ``Loihi: A neuromorphic manycore processor with on-chip learning,'' \textit{IEEE Micro}, vol.~38, no.~1, pp.~82-99, 2018.
\bibitem{Cauwenberghs13}
G. Cauwenberghs, ``Reverse engineering the cognitive brain,'' \textit{Proceedings of the National Academy of Sciences}, vol.~110, no.~39, pp.~15512-15513, 2013.
\vspace*{0.3mm}
\bibitem{Indiveri15}
G. Indiveri and S.-C. Liu, ``Memory and information processing in neuromorphic systems,'' \textit{Proceedings of the IEEE}, vol.~103, no.~8, pp.~1379-1397, 2015.
\bibitem{Park19}
J. Park, J. Lee and D. Jeon, ``A 65nm 236.5 nJ/classification neuromorphic processor with 7.5\% energy overhead on-chip learning using direct spike-only feedback,'' \textit{IEEE International Solid-State Circuits Conference-(ISSCC)}, pp.~140-142, 2019.
\bibitem{Kim15}
J. K. Kim et al., ``A 640M pixel/s 3.65 mW sparse event-driven neuromorphic object recognition processor with on-chip learning,'' \textit{IEEE Symposium on VLSI Circuits (VLSI-C)}, pp.~C50-C51, 2015.
\bibitem{Buhler17}
F. N. Buhler et al., ``A 3.43 TOPS/W 48.9 pJ/pixel 50.1 nJ/classification 512 analog neuron sparse coding neural network with on-chip learning and classification in 40nm CMOS,'' \textit{IEEE Symposium on VLSI Circuits (VLSI-C)}, pp.~C30-C31, 2017.
\bibitem{Chen18}
G. K. Chen et al., ``A 4096-neuron 1M-synapse 3.8pJ/SOP spiking neural network with on-chip STDP learning and sparse weights in 10nm FinFET CMOS,'' \textit{Proc. of IEEE Symp. on VLSI Circuits (VLSI-C)}, 2018.
\newpage%
\bibitem{Orchard15a}
G. Orchard et al., ``HFirst: A temporal approach to object recognition," \textit{IEEE Transactions on Pattern Analysis and Machine Intelligence}, vol.~37, no.~10, pp.~2028-2040, 2015.
\bibitem{Goetschalckx19}
K. Goetschalckx and M. Verhelst, ``Breaking High Resolution CNN Bandwidth Barriers with Enhanced Depth-First Execution,'' \textit{IEEE Journal on Emerging and Selected Topics in Circuits and Systems}, vol.~9, no.~2, pp.~323-331, 2019
\bibitem{Frenkel19c}
C. Frenkel, M. Lefebvre and D. Bol, ``Learning without feedback: Direct random target projection as a feedback-alignment algorithm with layerwise feedforward training,'' \textit{arXiv preprint arXiv:1909.01311}, 2019.
\bibitem{Rumelhart86}
D. Rumelhart, G. Hinton and R. Williams, ``Learning representations by back-propagating errors,'' \textit{Nature}, vol.~323, pp.~533-536, 1986.
\bibitem{Boahen00}
K. A. Boahen, ``Point-to-point connectivity between neuromorphic chips using address events,'' \textit{IEEE Transactions on Circuits and Systems II}, vol.~47, no.~5, pp.~416-434, 2000.
\bibitem{Vanarse16}
A. Vanarse, A. Osseiran and A. Rassau, ``A review of current neuromorphic approaches for vision, auditory, and olfactory sensors,'' \textit{Frontiers in Neuroscience}, no.~10, p.~115, 2016.
\bibitem{Thorpe01}
S. Thorpe, A. Delorme and R. Van Rullen, ``Spike-based strategies for rapid processing,'' \textit{Neural Networks}, vol.~14, no.~6-7, pp.~715-725, 2001.
\bibitem{Lillicrap16}
T. P. Lillicrap et al., ``Random synaptic feedback weights support error backpropagation for deep learning,'' \textit{Nature Communications}, vol.~7, no.~13276, 2016.
\bibitem{Nokland16}
A. N{\o}kland, ``Direct feedback alignment provides learning in deep neural networks,'' \textit{Proc. of Advances in Neural Information Processing Systems (NIPS)}, pp.~1037-1045, 2016.
\bibitem{Parhi99}
K. K. Parhi, \textit{VLSI Digital Signal Processing Systems: Design and Implementation}, John Wiley \& Sons, 1999.
\bibitem{LeCun98}
Y. LeCun and C. Cortes, ``The MNIST database of handwritten digits,'' 1998 [Online]. Available: \url{http://yann.lecun.com/exdb/mnist/}.
\bibitem{Whatmough17}
P. N. Whatmough et al., ``A 28nm SoC with a 1.2 GHz 568nJ/prediction sparse deep-neural-network engine with $>$0.1 timing error rate tolerance for IoT applications,'' \textit{Proc. of IEEE International Solid-State Circuits Conference (ISSCC)}, 2017.
\bibitem{Moons18}
B. Moons et al., ``BinarEye: An always-on energy-accuracy-scalable binary CNN processor with all memory on chip in 28nm CMOS,'' \textit{Proc. of IEEE Custom Integrated Circuits Conference (CICC)}, 2018.
\bibitem{Chen19}
Y. Chen et al., ``A 2.86-TOPS/W current mirror cross-bar-based machine-learning and physical unclonable function engine for Internet-of-Things applications" \textit{IEEE Transactions on Circuits and Systems I}, vol.~66, no.~6, pp.~2240-2252, 2019.
\bibitem{Mistry17}
K. Mistry, ``10nm technology leadership,'' \textit{Leading at the Edge: Intel Technology and Manufacturing Day,} 2017 [Online]. Available: \url{https://newsroom.intel.com/newsroom/wp-content/uploads/sites/11/2017/03/Kaizad-Mistry-2017-Manufacturing.pdf}.
\bibitem{Esser15}
S. K. Esser et al., ``Backpropagation for energy-efficient neuromorphic computing,'' \textit{Advances in Neural Information Processing Systems (NIPS}, pp.~1117-1125, 2015.
\bibitem{Paszke17}
A. Paszke et al., ``Automatic differentiation in PyTorch'', \textit{Proc. of Annual Conf. on Neural Information Processing Systems (NIPS) Workshop}, 2017.
\bibitem{Hubara17}
I. Hubara et al., ``Quantized neural networks: Training neural networks with low precision weights and activations,'' \textit{The Journal of Machine Learning Research}, vol.~18, no.~1, pp.~6869-6898, 2017.
\bibitem{Orchard15b}
G. Orchard et al., ``Converting static image datasets to spiking neuromorphic datasets using saccades,'' \textit{Frontiers in Neuroscience}, no.~9, p.~437, 2015.
\bibitem{Posch11}
C. Posch, D. Matolin and R. Wohlgenannt, ``A QVGA 143 dB dynamic range frame-free PWM image sensor With lossless pixel-level video compression and time-domain CDS,'' \textit{IEEE Journal of Solid State Circuits}, vol.~46, no.~1, pp.~259–275, 2011.
\end{thebibliography}
\end{document}